\documentclass[sigconf,nonacm]{acmart}

\AtBeginDocument{%
  \providecommand\BibTeX{{%
    \normalfont B\kern-0.5em{\scshape i\kern-0.25em b}\kern-0.8em\TeX}}}

\acmConference[]{}{}{}
\usepackage{graphicx}
\usepackage{color}
\usepackage{caption}
\usepackage{subcaption}
\usepackage{enumitem}
\usepackage{caption}
\usepackage{newfloat}
\usepackage{array}

\newcommand{\allprinciples}{\ref{item:relevant}--\ref{item:free}}

\newcommand{\youtubeTotal}{over~10K}

\begin{document}

\title[ML+X Tutorials]{Democratizing Machine Learning for Interdisciplinary Scholars: \\ Report on Organizing the NLP+CSS Online Tutorial Series}

\author{Ian Stewart}
\authornote{Both authors contributed equally to this research.}

\affiliation{
  \institution{Pacific Northwest National Laboratory}
  \country{}
}
\email{ian.stewart@pnnl.gov}

\author{Katherine A.~Keith}
\authornotemark[1]
\affiliation{%
  \institution{Williams College}
  \country{}
}
\email{kak5@williams.edu}

\renewcommand{\shortauthors}{}

\begin{abstract}
Many scientific fields---including biology, health, education, and the social sciences---use machine learning (ML) to help them analyze data at an unprecedented scale.
However, ML researchers who develop advanced methods rarely provide detailed tutorials showing how to apply these methods. Existing tutorials are often costly to participants, presume extensive programming knowledge, and are not tailored to specific application fields.
In an attempt to democratize ML methods, we organized a year-long, free, online tutorial series targeted at teaching advanced natural language processing (NLP) methods to computational social science (CSS) scholars.
Two organizers worked with fifteen subject matter experts to develop one-hour presentations with hands-on Python code for a range of ML methods and use cases, from data pre-processing to analyzing temporal variation of language change.
Although live participation was more limited than expected, a comparison of pre- and post-tutorial surveys showed an increase in participants' perceived knowledge of almost one point on a 7-point Likert scale.
Furthermore, participants asked thoughtful questions during tutorials and engaged readily with tutorial content afterwards, as demonstrated by \youtubeTotal~total views of posted tutorial recordings.
In this report, we summarize our organizational efforts and distill five principles for democratizing ML+X tutorials.
We hope future organizers improve upon these principles and continue to lower barriers to developing ML skills for researchers of all fields.\footnote{NLP+CSS Tutorial Website: \url{https://nlp-css-201-tutorials.github.io/nlp-css-201-tutorials/}}
\end{abstract}




\maketitle

\section{Introduction}\label{sec:intro}



Interest in incorporating machine learning into scientific analyses has exploded in the last two decades. 
Machine learning (ML)---the process of teaching a machine to predict statistical patterns in data \cite{jordan2015machine}---has gained prominence in biology \cite{jones2019setting}, physics \cite{karniadakis2021physics}, health care \cite{beam2018big}, and the social sciences \cite{mason2014computational} inter alia, yielding many successful ``ML+X'' collaborations.
While this potential for ML+X is enormous, many researchers unfamiliar with ML methods face barriers to entry, partly because implementing complex methods can be challenging for those without strong mathematical or programming backgrounds~\cite{cai2019software}.

\begin{figure}[t]
\centering
\includegraphics[width=0.8\columnwidth]{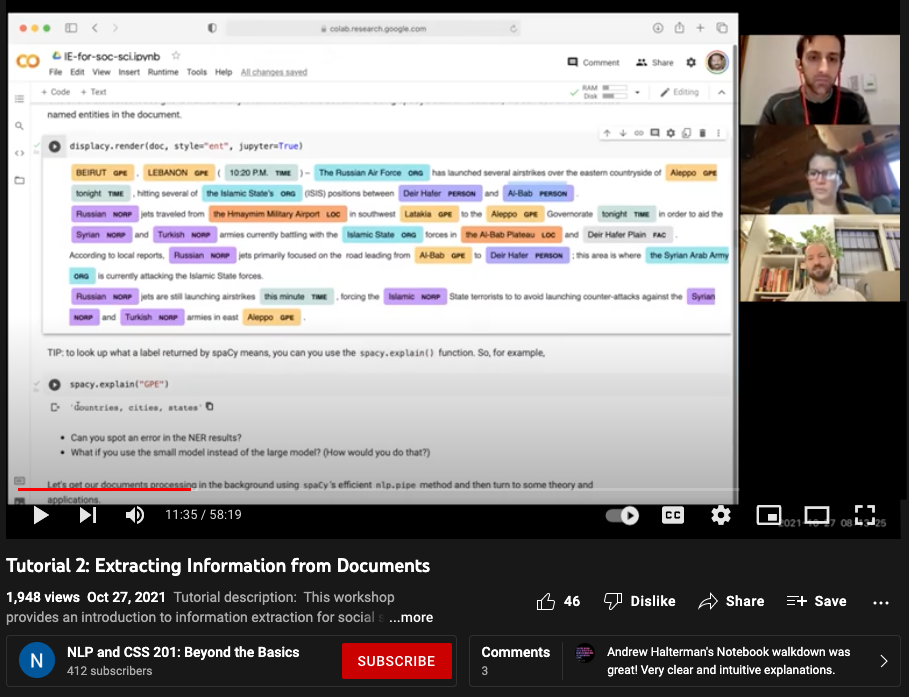}
\caption{Recording for \emph{Tutorial 2: Extracting Information from Documents} led by Andrew Halterman. As of October 23, 2022, this video had 5.1K views on YouTube. 
}
\label{f:andy}
\end{figure}

As a starting point for ML newcomers, some ML+X educational material covers simpler methods such as regression \cite{saunders2018eleven}.
For example, the Summer Institutes for Computational Social Science (SICSS) have developed learning materials that focus on basic ML methods for network analysis and text processing~\cite{sharlach2019}.
Computational social science researchers often leverage these kinds of methods to analyze web-scale data, which can shed light on complicated social processes such as political polarization~\cite{bail2018exposure}.

Basic ML methods provide a useful starting place but often lack the analytical power required to handle more complex questions that other fields require.
Social scientists often want to use ML to develop deep semantic representations of language or to estimate causal effects that can lead to better predictive performance.
Scientists who seek to advance their understanding of ML beyond the basics are often left searching for tutorial-like materials on their own, a difficult and often time-consuming task. 
On the other hand, well-meaning ML experts may try to share their expertise through media such as blog posts, but they run the risk of ``parachuting'' into unfamiliar fields with ill-adapted solutions~\cite{adame2021meaningful,summers2021artificial}.
Finally, many formal avenues for sharing knowledge about ML---such as academic conferences---can systematically \emph{exclude} researchers outside of ML via high fees to access materials.\footnote{Among other issues, social science research generally receives less funding compared to computer science. For instance, in 2021, the NSF dispersed \$283 million in funding for social, behavioral and economic sciences, versus \$1 billion for computer and information sciences and engineering (from \url{https://www.nsf.gov/about/congress/118/highlights/cu21.jsp}, accessed 10 August 2022). This lack of funding can often prevent social science researchers from attending ML conferences where many tutorials are presented.}

We take the position that ML researchers can make their methods more accessible and inclusive to researchers outside the field by creating online instruction explicitly tailored to the fields of subject matter experts.
Using the NLP+CSS tutorial series we organized in 2021-2022 as a case study, we argue that these interdisciplinary training sessions should incorporate the following \textbf{Principles for Democratizing ML+X Tutorials}:
\begin{enumerate}[label=\textbf{P.\arabic*}]
    \item Teach machine learning (ML) methods that are relevant and targeted to specific non-ML fields---e.g. biology, health, or the social sciences \label{item:relevant}
    \item Teach ML methods that are recent and cutting-edge  \label{item:recent}
    \item Lower start-up costs of programming languages and tooling\label{item:tooling}
    \item Provide open-source code that is clearly written, in context, and easily adapted to new problems \label{item:code}
    \item Reduce both monetary and time costs for participants \label{item:free}
\end{enumerate}



\paragraph{ML+Social Sciences}
Starting in summer 2021, we put our principles into action and created the \emph{NLP+CSS 201 Online Tutorial Series}.
We focused on an applied branch of machine learning to language data---a field called natural language processing (NLP)---aimed at early career researchers in the social sciences.
This report reflects on our experience and provides clear takeaways so that others can generalize our NLP + social sciences tutorials to tutorials targeted at other ML+X disciplines.


As we describe in Section~\ref{sec:methods}, we incorporated the principles above into our tutorial series by: (\ref{item:relevant}\&\ref{item:recent}) inviting experts in computational social science (CSS) to each lead a tutorial on a cutting edge NLP method; (\ref{item:tooling}\&\ref{item:code}) working with the experts to create a learning experience that is hosted in a self-contained interactive development environment in Python---Google CoLaboratory---and uses real-world social science datasets to provide context for the method; and (\ref{item:free}) hosting our tutorials live via Zoom and posting the recordings on YouTube, while providing all the materials and participation without any monetary costs to participants.

The impact of the inaugural year of our series is tangible.
We created twelve stand-alone tutorials made by fifteen area-expert tutorial hosts, have 396 members on an e-mail mailing list, and accumulated \youtubeTotal~total views on tutorial recordings posted to YouTube.\footnote{As of October 2022, videos available here: \url{https://www.youtube.com/channel/UCcFcF9DkanjaK3HEk7bsd-A}}
Comparing surveys pre- and post-tutorial, participants during the live sessions self-assessed as improving their knowledge of the topic by 0.77 on a 7-point Likert scale (Section~\ref{sec:analysis_effectiveness}).
After exploring highlights of the series, we discuss areas for improvement in Section~\ref{sec:conclusion}, including a suggestion to frame the tutorials a ``springboard'' for researcher's own exploration of advanced ML methods.


\section{Related work}

\subsection{Interdisciplinary tutorials}

Researchers specializing in NLP methods have proposed a variety of interdisciplinary tutorials to address social science questions, which we surveyed before we began planning our tutorial series. However, none satisfied all the principles we listed in Section~\ref{sec:intro}.
The tutorials presented at the conferences for the Association for Computational Linguistics (ACL)\footnote{\url{https://www.aclweb.org/portal/acl_sponsored_events}}---one of the premiere venues for NLP research---are on the cutting edge of research ($+$\ref{item:recent}) and often include code ($+$\ref{item:code}), but the ACL tutorials are also often are geared towards NLP researchers rather than researchers in fields outside of computer science ($-$\ref{item:relevant}), contain code that assumes substantial background knowledge ($-$\ref{item:tooling}) and cost hundreds of dollars to attend ($-$\ref{item:free}). Other interdisciplinary conferences such as the International Conference on Computational Social Science (IC2S2)\footnote{\url{https://iscss.org/ic2s2/conference/}} also have tutorials that explain recent NLP methods to computational social scientists ($+$\ref{item:relevant},\ref{item:recent},\ref{item:code}), but often the tutorials are presented with inconsistent formats ($-$\ref{item:tooling}) and cost money to attend ($-$\ref{item:free}). The Summer Institutes in Computational Social Science (SICSS) \cite{sharlach2019} provide free ($+$\ref{item:free}) tutorials on NLP methods for social scientists ($+$\ref{item:relevant}) with accompanying code ($+$\ref{item:tooling}\&\ref{item:code}), but they cover only the basic NLP techniques and not cutting edge methods ($-$\ref{item:recent}), while also limiting their target audience to people already involved with CSS research.\footnote{NLP methods include word counting and basic topic modeling: \url{https://sicss.io/curriculum} (accessed 11 August 2022).}


\subsection{Online learning}

While not without flaws, online learning experiences such as Massive Online Open Courses (MOOCs) have proven useful in higher education when meeting physically is impossible or impractical to due to students' geographic distance~\cite{de2011using,harasim2000shift,marcelino2018learning}.
Online courses have disrupted traditional education such as in-person college classes~\cite{vardi2012will}, but they may eventually prove most useful as a supplement rather than a replacement to traditional education~\cite{twigg2003models}.
For one, computer science students have found online learning useful when it incorporates interactive components such as hands-on exercises which may not be possible to execute during a lecture~\cite{meerbaum2013learning,tang2014environment}.
Additionally, while the centralized approach to traditional education can provide useful structure for students new to a domain, the decentralized approach of many online courses can provide room for socialization and creativity in content delivery~\cite{wallace2013social,wiley2002online}.
We intended our tutorial series to fit into the developing paradigm of online education as a decentralized and interactive experience, which would not replace but supplement social science education in machine learning. However, our tutorial series differs from MOOCs in that we limit the time committment for each topic to one hour (+\ref{item:free}) and each tutorial hour is meant to be stand-alone so that researchers can watch only the topics that are relevant to them. 


\section{Methods for Tutorial Series: Process and Timeline}\label{sec:methods}

\begin{table}[t]
  \centering
      \begin{tabular}{l p{6cm} >{\raggedleft \arraybackslash{}}p{1cm}}
      \toprule
      No.&Tutorial Title & Views \\
      \midrule
      \multicolumn{3}{c}{Fall 2021} \\
      T1 & Comparing Word Embedding Models & 1427 \\
      T2 & Extracting Information from Documents & 5386 \\
      T3 & Controlling for Text in Causal Inference with Double Machine Learning & 472 \\
      T4 & Text Analysis with Contextualized Topic Models & 570 \\
      T5 & BERT for Computational Social Scientists & 948 \\ \midrule
      \multicolumn{3}{c}{Spring 2022} \\       
      T6 & Moving from Words to Phrases when Doing NLP & 356 \\
      T7 & Analyzing Conversations in Python Using ConvoKit & 430 \\
      T8 & Preprocessing Social Media Text & 659 \\
      T9 & Aggregated Classification Pipelines & 139 \\
      T10 & Estimating Causal Effects of Aspects of Language with Noisy Proxies & 264 \\
      T11 & Processing Code-mixed Text & 259 \\
      T12 & Word Embeddings for Descriptive Corpus Analysis & 192 \\
      \bottomrule
  \end{tabular}
  \caption{Tutorial content. Order, title, and number of views of the corresponding recordings on YouTube as of October, 2022. Full abstracts of each tutorial are provided in the appendix, Table~\ref{tab:tutorial_abstracts}. 
  \label{t:tutorial-content}}
\end{table}


We describe our process and timeline for creating the tutorial series with the hope that future ML+X tutorial series organizers can copy or build from our experience. Throughout our planning process, we based our decisions on the five principles mentioned earlier (\ref{item:relevant}-\ref{item:free}). 
Our tutorial series spanned two semesters: Fall 2021 (August through December) and Spring 2022 (February through May).
The tutorial content is summarized in \autoref{t:tutorial-content}.

\subsection{Interest survey}
To identify relevant methods (\ref{item:relevant}), for one month before each semester we distributed a survey via our personal Twitter accounts, via a Google group mailing list that we created at the beginning of the fall 2021 semester, and via topically related mailing lists (e.g. a political methods list-serv).
We asked participants to list the methods that they would be most interested in learning about during a tutorial, which we then grouped into categories based on underlying similarities.

\begin{figure}
    \centering
    \includegraphics[width=\columnwidth]{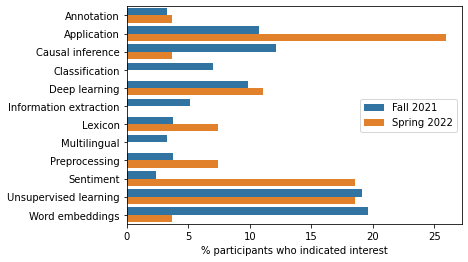}
    \caption{Distribution of NLP methods indicated in initial surveys.}
    \label{fig:interest_survey_distribution}
\end{figure}

The distribution of interest categories is shown in \autoref{fig:interest_survey_distribution}.
As expected, the responses covered many different NLP applications~\cite{nguyen2020we}, including data preparation (preprocessing, multilingual), conversion of text to relevant constructs (information extraction, word embeddings, deep learning), and downstream analysis (causal inference, application).
Most participants expressed interest in word embeddings, unsupervised learning, and downstream applications of NLP methods, which aligns with the current popularity of such methods.

\paragraph{Lessons learned}
Since we typically publish in NLP venues, we took a prescriptive approach to choosing the tutorial methods to present, in an attempt to more actively shape the field of computational social science (addressing \ref{item:relevant}\& \ref{item:recent}).
We used the results of the survey to brainstorm potential topics for each upcoming semester, but did not restrict ourselves to only the most popular methods.
While useful, the interest surveys revealed a disconnect between our ideal tutorials, which focused on advanced NLP methods, and the participants' ideal tutorials, e.g. entry-level methods with immediate downstream results.
For example, many participants in the Spring 2022 interest survey mentioned sentiment analysis, a well-studied area of NLP~\cite{birjali2021} that we considered to be more introductory-level and sometimes unreliable~\cite{diaz2018addressing}.
This was one source of tension between our expectations and those of the participants, and future tutorial series organizers may want to focus their efforts on highly-requested topics to ensure consistent participation and satisfaction (\ref{item:relevant}).

\subsection{Leader recruitment}
Aligning with \ref{item:recent}, we recruited other NLP experts who worked on cutting-edge methods to lead each individual tutorial.\footnote{We recruited tutorial leaders through our own social networks and through mutual acquaintances. We targeted post-doctoral fellows, early-career professors, and advanced graduate students.} 
To ensure \ref{item:tooling}, we also met with the tutorial hosts to agree on a common format for the programming platform---Google CoLaboratory with Python\footnote{\url{https://colab.research.google.com/}}---and to help them understand the tutorials' objectives.
The process involved several meetings: an introduction meeting to scope the tutorial, and at least one planning meeting to review the slides and code to be presented.
Normally, this process was guided by a paper or project for which the tutorial leader had code available.
For example, the leader of tutorial T4 was able to leverage an extensive code base already tested by her lab.

\paragraph{Lessons learned}
During the planning process, we were forced to plan the tutorials one at a time due to complicated schedules among the leaders. 
We spread out the planning meetings during the semester so that the planning meetings would begin roughly two to three weeks before the associated tutorial.
We strongly encouraged leaders to provide their code to us at least one week in advance to give us time to review it, but we found this difficult to enforce due to time constraints on the leaders' side (e.g. some leaders had to prioritize other teaching commitments).
Future organizers should set up a consistent schedule for contacting leaders in advance and agree with leaders on tutorial code that is relatively new and usable (\ref{item:recent}\& \ref{item:code}) without presenting an undue burden for the leader, e.g. re-using existing code bases.

\subsection{Participant recruitment}
Even if we guaranteed \ref{item:relevant}-\ref{item:free} with the content developed, recruiting social science participants was essential to the success of our tutorial series.
In September 2021, we set up an official mailing list through Google Groups and advertised it on social media and other methods-related list-servs.\footnote{The Google Group was only accessible to participants with Google Mail accounts, which in retrospect likely discouraged some participants who only use institutional email accounts.}
The mailing list eventually hosted 396 unique participants.
For all tutorials, we set up a RSVP system using Google Forms for participants to sign up, and we provided an RSVP link up to one week before each tutorial.
We chose this ``walled garden'' approach to discourage anti-social activity such as Zoom-bombing which is often made easier by open invitation links~\cite{ling2021first}, and to provide tutorial leaders with a better sense of their participants.

\begin{figure}
    \centering
    \begin{subfigure}{\columnwidth}
      \includegraphics[width=\columnwidth]{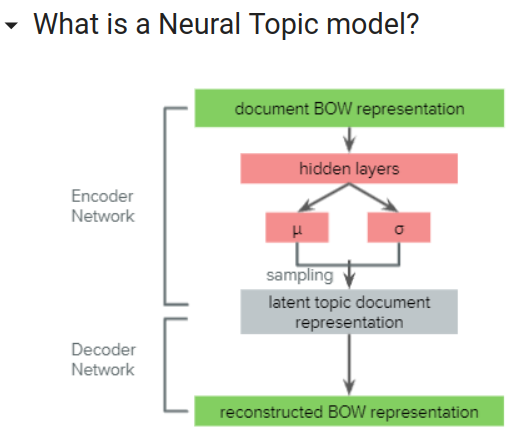}
      \caption{\label{fig:topic_model_diagram}}
    \end{subfigure}
    \begin{subfigure}{\columnwidth}
      \includegraphics[width=\textwidth]{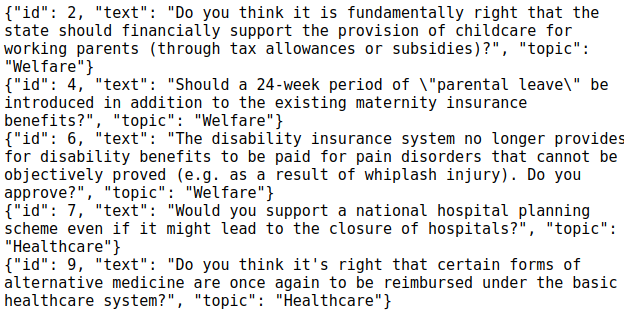}
      \caption{\label{fig:topic_model_docs}}
    \end{subfigure}
    \begin{subfigure}{\columnwidth}
      \includegraphics[width=\columnwidth]{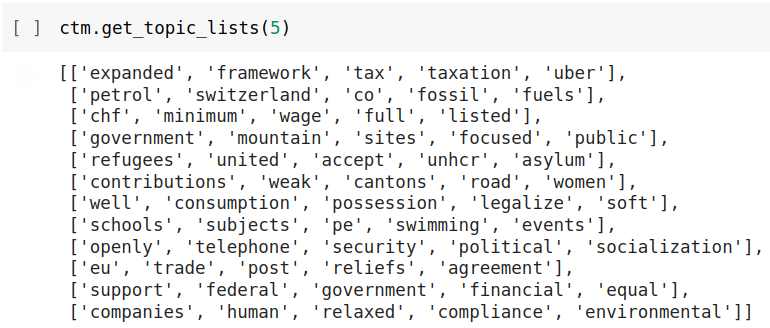}
      \caption{\label{fig:topic_model_words}}
    \end{subfigure}
    \caption{Excerpts from the tutorial on topic modeling (T4), demonstrating (a) the application of a neural model to (b) text from politicians' interviews, which produces (c) word lists for inductively discovered topics.}
    \label{fig:tutorial_analysis_example}
\end{figure}

\paragraph{Lesson learned}
This process revealed significant drop-out: between 10-30\% of people who signed up actually attended the tutorial.
While the reasons for the drop-out remained unclear, we reasoned that people signed up for the tutorial as a back-up and were willing to miss the live session if another obligation arose, under the assumption that the recording would be available later. Although we believe in the benefits of asynchronous learning, the low number of live participants was somewhat discouraging to the live tutorial hosts.

\subsection{Running the tutorials}
During the tutorials, we wanted to ensure low start-up cost of the programming environment (\ref{item:tooling}) and well-written code that participants could use immediately after the tutorials (\ref{item:code}).
We designed each tutorial to run for slightly under 60 minutes, to account for time required for introductions, transitions, and follow-up questions.
The tutorial leader began the session with a presentation to explain the method of interest with minimal math, using worked examples on toy data and examples of prior research that leveraged the method.

After 20-30 minutes of presentation, the tutorial leader switched to showing the code written in a Google CoLaboratory Python notebook (\ref{item:tooling}), which is an internet-based coding environment that allows users to run modular blocks of Python code.
The leader would load or generate a simple text dataset, often no more than several hundred documents in size, to illustrate the method's application.
Depending on the complexity of the method, the leader might start with some basic steps and then show the students increasingly complicated code snippets.
In general, the leaders walked the students through separate modules that showed different aspects of the method in question.
During the topic modeling session (T4), the leader showed first how to train the topic model, then provided extensive examples of what the topic output looked like and how it should be interpreted (e.g. top words per topic, example documents with high topic probabilities).\footnote{Topic models are used to identify latent groupings for words in a document, e.g. a health-related topic might include ``exercise'' and ``nutrition.''~\cite{blei2003latent}}
As a point of comparison, the leader also would often show the output of a simpler ``baseline'' model
to demonstrate the superior performance of the tutorial's more advanced method.
We show excerpts from the tutorial notebook on topic modeling in \autoref{fig:tutorial_analysis_example}, which includes an overview of the topic model, a sample of the text data which relates to politics, and the resulting learned ``topics'' as lists of words.


\paragraph{Lessons learned}

To encourage critical thinking, some of the tutorial leaders provided questions or exercises in the Colab notebooks for students to complete at a later time.
The leader of the information extraction tutorial (T2) created an exercise for students to parse sentences from news text related to military activity, and then to extract all sentences that described an attack between armies.
Some of these exercises posed challenges to participants who lacked experience with the data structures or function calls involved in the code.
For future tutorials, leaders should consider simply showing participants how to solve a simple exercise (e.g. live-coding) rather than expecting participants to attack the problem on their own.

\begin{table*}[t!]
\centering
\begin{tabular}{l p{16.5cm}}
Tutorial & Question sample \\ \toprule
T1 &  Is there a reason that we're using word2vec rather than other models such as fastText?  What does Euclidean distance between embeddings mean? Does word2vec work on short ``documents'' such as Twitter data? \\
T4 & How is the bag of words representation combined with contextualized representation? How should someone choose the model to use for this component? What models does your package support? \\
T6 &  Are Phrases mostly Nouns since Nouns are the ones that have multi-words?  Have you tried this model in languages other than English? How was the PhraseBert model trained? \\
T7 &  How do you keep track of who's responding to what previous utterance?  How do you create a conversation corpus from scratch?  Can the code provide statistics or summary for each speaker or utterance? \\
T9 &  Could you interpret a well-calibrated model as estimating the moral outrage in a post? Do choices in favor of hard modeling an aggregation techniques lead to higher values in outcome measurement? What would you recommend to handle annotator disagreement when the task is to label spans inside the text? \\ \bottomrule
\end{tabular}
\caption{Example questions from tutorial sessions. Some wording changed for clarity.}
\label{tab:tutorial_example_questions}
\end{table*}

\begin{figure}[t!]
\centering
\begin{subfigure}{\columnwidth}
  \centering
  \includegraphics[width=\columnwidth]{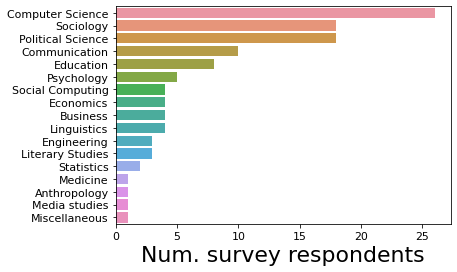}
  \caption{\label{f:fields}}
\end{subfigure}

\begin{subfigure}{.8\columnwidth}
  \centering
  \includegraphics[width=\columnwidth]{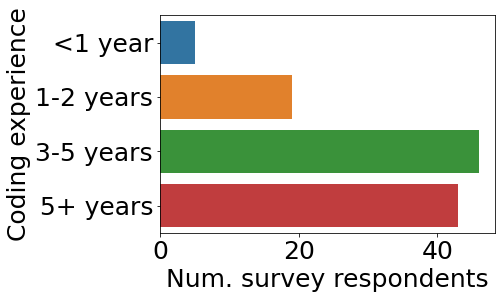}
  \caption{  \label{f:experience}}
\end{subfigure}
\begin{subfigure}{\columnwidth}
  \centering
  \resizebox{0.98\linewidth}{!}{ 
      \begin{tabular}{lrr}
      \toprule 
      \toprule
       & $\mu$ & $\sigma$  \\
      \toprule 
     \toprule
      \textbf{Pre-Q1}--Learned from code (1-5) & 4.00 & 0.94\\
    \textbf{Pre-Q2}--Learned from content (1-5)& 4.24 & 0.90 \\
      \bottomrule 
      \toprule 
      \textbf{Pre- vs post-survey} & E[\textbf{Post-Q3}] - E[\textbf{Pre-Q3}] \\
      \toprule 
      \toprule
      Knowledge about the topic (1-7) & $0.77^{*}$ \\
      \bottomrule 
      \bottomrule
  \end{tabular}
  }
  \caption{\label{t:post-survey-stats}}
\end{subfigure}

\caption{Participant responses for the survey sent during the live tutorials (aggregated from T4-T12). Figure (a) indicates participant disciplines \textbf{(Pre-Q1)} and (b) coding experience \textbf{(Pre-Q2)}. Table (c) shows \emph{(top)} the mean ($\mu)$ and standard deviation ($sigma$) on a 5-point Likert scale for \textbf{Post-Q1\&2} and \emph{(bottom)} for the question about participants' self-rated knowledge about the topic graded on a 7-point Likert scale, the expected value of the post survey minus the expected value of the pre-survey (E[\textbf{Post-Q3}] - E[\textbf{Pre-Q3}]). $^*$ indicates statistical significance with p-value $<10^{-5}$ via a two-sided T-test.}
\label{fig:allsurveys}
\end{figure}

\subsection{Participation during tutorials}

During each tutorial, we---the authors of the study---acted as facilitators to help the leaders handle questions and manage time effectively.
The leaders were often unable to see the live chat while presenting, and we therefore found natural break points in the presentation to answer questions sent to the chat.
While we allowed for written and spoken questions, participants preferred to ask questions in the chat, possibly to avoid interrupting the presenter and to allow them to answer asynchronously.

Participants were encouraged to test out the code on their own during the tutorial, and the code was generally written to execute quickly without significant lag for e.g. downloads or model training (\ref{item:tooling}).
This often required the leaders to run some of the code in advance to automate less interesting components of the tutorial, e.g. selecting the optimal number of topics for the topic model.

\paragraph{Lessons learned}
Based on some of the questions received, participants seemed to engage well with the code and to follow up with some of the methods.
Participants asked between 1 and 15 questions per tutorial (median 5).
We show example questions from the tutorials with the largest number of questions in \autoref{tab:tutorial_example_questions}.
The questions cover both simple closed-answer questions (``Can the code provide statistics'') and more complicated open-ended questions (``How should someone choose the model to use'').
While the number of questions was relatively low overall, the participants who asked questions were engaged and curious about the limitations and ramifications of the methods being presented.
To improve participant engagement via questions, future leaders may find it useful to pose their own questions throughout the code notebook (``what do you think would happen if we applied method X while setting parameter Z=1?'') as a way to guide the participants' curiosity.





\section{Analysis of Effectiveness}
\label{sec:analysis_effectiveness}

\subsection{Pre- and post-surveys during live tutorials} 
During the live portions of the tutorials, we distributed an optional survey to participants at the beginning and end of the one-hour sessions.\footnote{During T1-T3 we were prototyping the series, so we only distributed the surveys for T4-T12.} The pre-survey consisted of three questions in a Google form: \textbf{(Pre-Q1)} \emph{Academic discipline background} in which participants chose one of the given disciplines or wrote their own;
\textbf{(Pre-Q2)} \emph{How many years of experience in coding/data analysis do you have?} which had four options;
and \textbf{(Pre-Q3)} \emph{How much do you currently know about the topic?} which was judged on a 7-point Likert scale with 1 described as \emph{I know nothing about the topic}, 4 described as \emph{I could possibly use the methods in my research, but I'd need guidance} and 7 described as \emph{Knowledgeable, I could teach this tutorial.} The post-survey consisted of four questions: 
\textbf{(Post-Q1)} \emph{Code: How much did you learn from the hands-on code aspect of the tutorial?}; \textbf{(Post-Q2)}  \emph{Content: How much did you learn from the content part of the tutorial?};  \textbf{(Post-Q3)} \emph{Now, after the tutorial, how much do you currently know about the topic?} and \textbf{(Post-Q4)} \emph{Any suggestions or changes we should make for the next tutorial?}. Questions 1 and 2 were judged on a 5-point Likert scale with 1 described as \emph{Learned nothing new} and 5 described as \emph{Learned much more than I could have on my own}. Question 3 was judged on the same 7-point Likert scale as the analogous question in the pre-survey. 

\paragraph{Results}
We report aggregated survey responses in Figure~\ref{fig:allsurveys}. 
Across the eight tutorials for which we collected data, the pre-surveys had 113 respondents total and the post-surveys had 63 respondents. 
Figure~\ref{f:fields} shows the results of the breakdown by academic discipline or background \textbf{(Pre-Q1)}. The three largest areas of participation came from the fields of computer science, sociology, and political science. Figure~\ref{f:experience} shows that our participants actually had quite a lot of experience in coding or data analysis \textbf{(Pre-Q2)}--78.8\% of participants who responded had three or greater years of experience in coding.

Analyzing \textbf{Post-Q1} about how much they learned from code, participants responded with $\mu=4, \sigma=0.94$.
\textbf{Post-Q2} about learning from content was similar with $\mu=4.24, \sigma=0.9$. Interpreting these results, many participants perceived a high degree of learning from attending the live tutorials.
We measure the pre- to post-survey learning by computing the difference between the mean of \textbf{Post-Q3} and mean of \textbf{Pre-Q3}, and we find a difference of 0.77.\footnote{Ideally, we would look not at the aggregate participant responses but instead test the pairwise differences for each individual's pre- versus post-survey. However, we found that only 18 participants could be matched from pre- to post-survey due to drop-out, which is too small for pairwise significance testing.}
We ran a two-sided T-test to see if the pre- versus post-survey differences were greater than zero with statistical significance, which produced a t-value of 4.16 and a p-value less than $10^{-5}$.
While seemingly small in aggregate, this change represents a consistent growth in perceived knowledge among participants that is surprising considering the relatively short tutorial length of one hour.
Manually reading the responses from \textbf{(Post-Q4)}, participants described very positive experiences, including ''very good tutorial'' ``Excellent tutorial!!!'' and ``very helpful.'' 

\paragraph{Lessons learned}
As Figure~\ref{f:fields} shows, we were successful in recruiting participants from a wide variety of social science disciplines. However, computer science or data science---top-most bar in Figure~\ref{f:fields}---was the most represented field. In reflection, having another organizer who was primarily focused on social science, rather than NLP, would help us recruit more CSS-oriented participants and would align better with \ref{item:relevant}. 
Responses from \textbf{Post-Q4} also indicated that the tutorials were not long enough for some participants. One participant said ``It would be great to make something like this into a multi-part tutorial. It seemed like too much new material for 1 hour.'' Some suggestions for future tutorial organizers could be to make the tutorials 2-3 hours long. In the first hour, the tutorial could provide an overview, followed by more advanced topics or practice in hours 2-3. It's difficult to satisfy the trade-offs of (1) audience attention bandwidth and (2) fully explaining a particular method. We also could have improved how we set audience expectations: introducing the tutorials as a crash course and explaining that participants should expect to spend 4-5 hours on their own afterwards to learn the material in depth.
Furthermore, future leaders may want to require or strongly encourage participation in the surveys to improve data collection as we had relatively low participation rates.\footnote{After all the tutorials were presented, we also sent a survey to the mailing list to ask about how much participants had learned from the tutorials and whether they used the material in their own work.
We received only five responses total, therefore we do not present statistics here.
} 

\subsection{Downstream impact}
Despite the relatively low synchronous participation (roughly 4-30 participants per session), the views on the tutorial videos posted to YouTube showed consistent growth during the tutorial series and even afterward, culminating in \youtubeTotal~total views.
In addition, the tutorial materials were showcased on the website for the Summer Institute for Computational Social Science,\footnote{Accessed 15 October 2022: \url{https://sicss.io/overview}.} and several tutorial leaders presented their tutorials again at an international social science conference, having prepared relevant materials as part of our series (\ref{item:code}). \footnote{International Conference on Web and Social Media 2022, accessed 15 October 2022: \url{https://www.icwsm.org/2022/index.html/\#tutorials-schedule}}
The tutorial series may therefore have the greatest impact not for the synchronous participants but instead for the large and growing audience of researchers who discover the materials after the fact and may not have the resources to learn about the methods via traditional methods (\ref{item:free}).
The success of the tutorials in other contexts also points to the beginning of a virtuous cycle, in which tutorial leaders test-drive their work in an informal setting and then present a more formal version at an academic conference.

\section{Conclusion}\label{sec:conclusion}



\paragraph{Future improvements}
Reflecting on this experience report, we suggest the following improvements for future ML+X tutorial organizers:


\begin{itemize}[leftmargin=*]
    \item Despite the results of the pre-tutorial interest surveys, we made curatorial decisions about the content and we cannot be sure that we satisfied the needs of what participants wanted versus what we thought was important.
    Future organizers may achieve higher participation by focusing on methods with high public interest, regardless of their lower perceived utility by subject area experts.
    \item The two co-organizers were both computer scientists, and we largely leveraged a computer science professional network for recruitment. 
    Future renditions would ideally include a social scientist co-organizer to provide better insight into current ML needs and desires among researchers (\ref{item:relevant}), as well as helping tutorial participants feel more at ease with complicated ML methods.
     \item Despite high sign-up rates and lack of cost (\ref{item:free}), participants would often fail to attend tutorials for which they had signed up. 
    This may reflect a lack of commitment among participants due to the virtual presence (``just another Zoom meeting'')~\cite{toney2021fighting}, or a failure to send frequent reminders.
    Future tutorial organizers should experiment with other strategies for encouraging attendance, including more topical data sets, a ``hackathon'' setting~\cite{mtsweni2015stimulating}, or a structured community to engage participants before and after the tutorial~\cite{harasim2000shift}.
    \item We found that participants did not consistently engage in the hands-on coding segments of the tutorials.
    We recommend that future tutorial leaders either simplify the hands-on coding for short sessions, or follow up on the tutorial with additional ``office hours'' for interested students to try out the code and ask further questions about the method.
    Similar to some computer science courses, this approach might have a lecture component and a separate ``recitation'' session for asking questions about the code.
    \item In the early stages of the tutorial series, we focused more on executing the tutorials rather than collecting quantitative data about the participants' experience. 
    This makes it difficult to judge some aspects of the tutorials' success, especially how the tutorials were received by participants with different backgrounds and expectations.
    With more extensive evaluation and participation in surveys, we hope that future organizers will make quicker and more effective improvements during the course of a tutorial series. 
\end{itemize}

\paragraph{Successes}
Despite these drawbacks, we believe our tutorial series succeeded in its goal---to help social scientists advance their skills beyond introductory NLP methods. 
We hope other ML+X tutorials can build from our successes: 
\begin{itemize}[leftmargin=*]
\item We accumulated \youtubeTotal~total views among our public recordings.
    Thus, we'd encourage future ML+X organizers to put even more effort into the recordings rather than live sessions. 
    \item 
     Although participants came in skilled---78.8\% of participants who responded had three or greater years of experience in coding (Figure~\ref{f:experience})--they reported aggregate increase in perceived knowledge of the methods presented---0.77 on a 7-point Likert scale. 
    \item We generated education content for a diverse set of relevant and new NLP methods (\ref{item:relevant}\&\ref{item:recent}) that can accelerate social science research. 
    The subject matter experts who led the tutorials were able to translate complicated ML concepts into understandable, step-by-step lessons.
    We hope future ML+X organizers can take inspiration from these tutorials' choice of content and social organization.
    \item Our tutorials have produced ready-to-use, modular, and freely available Python code with a low barrier to entry (\ref{item:tooling},\ref{item:code},\ref{item:free}), which will provide ``scaffolding'' to future students seeking to start their own projects~\cite{nam2010effects}.
    We envision future ML+X organizers using this codebase as a template for releasing code in their own domain. 
\end{itemize}




As machine learning methods become more available and more powerful, scientists may feel encouraged to implement these methods within their own domain-specific research. 
We believe tutorial series such as the one described in this report will help guide these researchers on their journey. 
Like the tutorials themselves, we hope that our \emph{Principles for Democratizing ML+X Tutorials} (\allprinciples) will be used as springboard toward more open and inclusive learning experiences for all researchers.
Rather than wait for top-down solutions, we encourage other ML practitioners to get involved and shape the future of applied science by sharing their knowledge directly with scholars eager to know more.

\section*{Acknowledgments}
We are deeply grateful for financial assistance from a Social Science Research Council (SSRC)/Summer Institutes in Computational Social Science (SICSS) Research Grant.
We thank SIGCSE reviewers and various computer science education experts for their feedback on initial drafts.
We thank the fifteen organizers who generously donated their expertise and time to making these tutorials possible: Connor Gilroy, Sandeep Soni, Andrew Halterman, Emaad Manzoor, Silvia Terragni, Maria Antoniak, Abe Handler, Shufan Wang, Jonathan Chang, Steve Wilson, Dhanya Sridhar, Monojit Choudhury, Sanad Rizvi, and Neha Kennard.

\bibliographystyle{acm}
\bibliography{bib}

\newpage
\section*{Appendix}

We provide the full abstracts of the tutorials in Table~\ref{tab:tutorial_abstracts}, which the tutorial leaders wrote in coordination with the organizers.

\begin{table*}[t!]
\small
\begin{tabular}{l p{17cm}}
~ & Summary \\ \toprule
T1 & We’ll demonstrate an extension of the use of word embedding models by fitting multiple models on a social science corpus (using gensim’s word2vec implementation), then aligning and comparing those models. This method is used to explore group variation and temporal change. We’ll discuss some tradeoffs and possible extensions of this approach. \\ \hline
T2 & This workshop provides an introduction to information extraction for social science–techniques for identifying specific words, phrases, or pieces of information contained within documents. It focuses on two common techniques, named entity recognition and dependency parses, and shows how they can provide useful descriptive data about the civil war in Syria. The workshop uses the Python library spaCy, but no previous experience is needed beyond familiarity with Python. \\ \hline
T3 & Establishing causal relationships is a fundamental goal of scientific research. Text plays an increasingly important role in the study of causal relationships across domains especially for observational (non-experimental) data. Specifically, text can serve as a valuable ``control'' to eliminate the effects of variables that threaten the validity of the causal inference process. But how does one control for text, an unstructured and nebulous quantity? In this tutorial, we will learn about bias from confounding, motivation for using text as a proxy for confounders, apply a ``double machine learning'' framework that uses text to remove confounding bias, and compare this framework with non-causal text dimensionality reduction alternatives such as topic modeling. \\ \hline
T4 & Most topic models still use Bag-Of-Words (BoW) document representations as input. These representations, though, disregard the syntactic and semantic relationships among the words in a document, the two main linguistic avenues to coherent text. Recently, pre-trained contextualized embeddings have enabled exciting new results in several NLP tasks, mapping a sentence to a vector representation. Contextualized Topic Models (CTM) combine contextualized embeddings with neural topic models to increase the quality of the topics. Moreover, using multilingual embeddings allows the model to learn topics in one language and predict them for documents in unseen languages, thus addressing a task of zero-shot cross-lingual topic modeling. \\ \hline
T5 & What is BERT? How do you use it? What kinds of computational social science projects would BERT be most useful for? Join for a conceptual overview of this popular natural language processing (NLP) model as well as a hands-on, code-based tutorial that demonstrates how to train and fine-tune a BERT model using HuggingFace’s popular Python library. \\ \hline
T6 & Most people starting out with NLP think of text in terms of single-word units called ``unigrams.'' But many concepts in documents can’t be represented by single words. For instance, the single words ``New'' and ``York'' can’t really represent the concept ``New York.'' In this tutorial, you’ll get hands-on practice using the phrasemachine package and the Phrase-BERT model to 1) extract multi-word expressions from a corpus of U.S. Supreme Court arguments and 2) use such phrases for downstream analysis tasks, such as analyzing the use of phrases among different groups or describing latent topics from a corpus. \\ \hline
T7 & ConvoKit is a Python toolkit for analyzing conversational data. It implements a number of conversational analysis methods and algorithms spanning from classical NLP techniques to the latest cutting edge, and also offers a database of conversational corpora in a standardized format. This tutorial will walk through an example of how to use ConvoKit, starting from loading a conversational corpus and building up to running several analyses and visualizations. \\ \hline
T8 & hmm howwww should we think about our \#NLProc preprocessing pipeline when it comes to informal TEXT written by social media users?!? In this tutorial, we’ll discuss some interesting features of social media text data and how we can think about handling them when doing computational text analyses. We will introduce some Python libraries and code that you can use to process text and give you a chance to experiment with some real data from platforms like Twitter and Reddit. \\ \hline
T9 & NLP has helped massively scale-up previously small-scale content analyses. Many social scientists train NLP classifiers and then measure social constructs (e.g sentiment) for millions of unlabeled documents which are then used as variables in downstream causal analyses. However, there are many points when one can make hard (non-probabilistic) or soft (probabilistic) assumptions in pipelines that use text classifiers: (a) adjudicating training labels from multiple annotators, (b) training supervised classifiers, and (c) aggregating individual-level classifications at inference time. In practice, propagating these hard versus soft choices down the pipeline can dramatically change the values of final social measurements. In this tutorial, we will walk through data and Python code of a real-world social science research pipeline that uses NLP classifiers to infer many users’ aggregate ``moral outrage'' expression on Twitter. Along the way, we will quantify the sensitivity of our pipeline to these hard versus soft choices. \\ \hline
T10 & Does the politeness of an email or a complaint affect how quickly someone responds to it? This question requires a causal inference: how quickly would someone have responded to an email had it not been polite? With observational data, causal inference requires ruling out all the other reasons why polite emails might be correlated with fast responses. To complicate matters, aspects of language such as politeness are not labeled in observed datasets. Instead, we typically use lexicons or trained classifiers to predict these properties for each text, creating a (probably noisy) proxy of the linguistic aspect of interest. In this talk, I’ll first review the challenges of causal inference from observational data. Then, I’ll use the motivating example of politeness and response times to highlight the specific challenges to causal inference introduced by working with text and noisy proxies. Next, I’ll introduce recent results that establish assumptions and a methodology under which valid causal inference is possible. Finally, I’ll demonstrate this methodology: we’ll use semi-synthetic data and adapt a text representation method to recover causal effect estimates. \\ \hline
T11 & Code-mixing, i.e., the mixing of two or more languages in a single utterance or conversation, is an extremely common phenomenon in multilingual societies. It is amply present in user-generated text, especially in social media. Therefore, CSS research that handles such text requires to process code-mixing; there are also interesting CSS and socio-linguistic questions around the phenomenon of code-mixing itself. In this tutorial, we will equip you with some basic tools and techniques for processing code-mixed text, starting with hands-on experiments with word-level language identification, all the way up to methods for building code-mixed text classifiers using massively multilingual language models. \\ \hline
T12 & Word embeddings such as word2vec have recently garnered attention as potentially useful tools for analysis in social science. They promise an unsupervised method to quantify the connotations of words, and compare these across time or different subgroups. However, when training or using word embeddings, researchers may find that they don’t work as well as expected, or produce unreplicable results. We focus on three subtle issues in their use that could result in misleading observations: (1) indiscriminate use of analogical reasoning, which has been shown to underperform on many types of analogies; (2) the surprising prevalence of polysemous words and distributional similarity of antonyms, both leading to counterintuitive results; and (3) instability in nearest-neighbor distances caused by sensitivity to noise in the training process. Through demonstrations, we will learn how to detect, understand, and most importantly mitigate the effects of these issues. \\ \bottomrule
\end{tabular}
\caption{Tutorial abstracts, provided by leaders.
\label{tab:tutorial_abstracts}}
\end{table*}


\end{document}